\documentclass[runningheads]{llncs}
\usepackage{graphicx}
\usepackage{multirow}
\usepackage{subcaption}
\usepackage{hyperref}

\usepackage{tikz}
\usepackage{comment}
\usepackage{amsmath,amssymb} % define this before the line numbering.
\usepackage{color}

\usepackage[width=122mm,left=12mm,paperwidth=146mm,height=193mm,top=12mm,paperheight=217mm]{geometry}
\usepackage[latin1]{inputenc}

\begin{document}
% \linenumbers
\pagestyle{headings}
\mainmatter
\def\ECCVSubNumber{13}  % Insert your submission number here

\title{Late Temporal Modeling in 3D CNN Architectures with BERT for Action Recognition} % Replace with your title

\begin{comment}
\titlerunning{ECCV-20 submission ID \ECCVSubNumber} 
\authorrunning{ECCV-20 submission ID \ECCVSubNumber} 
\author{Anonymous ECCV submission}
\institute{Paper ID \ECCVSubNumber}
\end{comment}
%******************

%\begin{comment}
\titlerunning{Late Temporal Modeling in 3D CNNs with BERT}
%\begin{comment}
\author{\index{Kalfaoglu, M. Esat}  \inst{1}\orcidID{0000-1111-2222-3333} \and
Sinan Kalkan\inst{2,3}\orcidID{1111-2222-3333-4444} \and
\index{Alatan, A. Aydin}  \inst{3}\orcidID{2222--3333-4444-5555}}
\authorrunning{E. Kalfaoglu et al.}
%\end{comment}
\author{M. Esat Kalfaoglu \inst{1,3}\orcidID{0000-0001-5942-0454}\and
Sinan Kalkan \inst{2,3}\orcidID{0000-0003-0915-5917}\and
A. Aydin Alatan \inst{1,3}\orcidID{0000-0001-5556-7301}}
\institute{Department of Electrical and Electronics Engineering \and
Department of Computer Engineering \and
Center for Image Analysis (OGAM) \\
Middle East Technical University, Ankara, Turkey 
\email{\{esat.kalfaoglu,skalkan,alatan\}@metu.edu.tr}
}
%\end{comment}
%******************
\maketitle

\begin{abstract}
In this work, we combine 3D convolution with late temporal modeling for action recognition. For this aim, we replace the conventional Temporal Global Average Pooling (TGAP) layer at the end of 3D convolutional architecture with the Bidirectional Encoder Representations from Transformers (BERT) layer in order to better utilize the temporal information with BERT's attention mechanism. We show that this replacement improves the performances of many popular 3D convolution architectures for action recognition, including ResNeXt, I3D, SlowFast and R(2+1)D. Moreover, we provide the-state-of-the-art results on both HMDB51 and UCF101 datasets with 85.10\% and 98.69\% top-1 accuracy, respectively. The code is publicly available\footnote{\href{https://github.com/artest08/LateTemporalModeling3DCNN}{\color{magenta} github.com/artest08/LateTemporalModeling3DCNN}}.
\end{abstract}

\section{Introduction}
Action Recognition (AR) pertains to identifying the label of the action or the activity observed in a video clip. With cameras everywhere, AR has become essential in many domains, such as video retrieval, surveillance, human-computer interaction, and robotics.

A video clip contains two critical pieces of information for AR: Spatial and temporal information. Spatial information represents the static information in the scene, such as objects, context, entities, etc., which are visible in a single frame of the video, whereas temporal information, obtained by integrating the spatial information over frames, mostly captures the dynamic nature of the action.

In this work, the joint utilization of two temporal modeling concepts from the literature, which are 3D convolution and late temporal modeling, is proposed and analyzed. Briefly, 3D convolution is a way of generating a temporal relationship hierarchically from the beginning to the end of CNN architectures. On the other hand, late temporal modeling is typically utilized with 2D CNN architectures, where the features extracted by 2D CNN architectures from the selected frames are usually modeled with recurrent architectures, such as LSTM, Conv LSTM. 

Despite its advantages, the temporal global average pooling (TGAP) layer which is used at the end of all 3D CNN architectures \cite{Carreira2017,Chen2018Multi-fiberRecognition,Feichtenhofer2019SlowfastRecognition,Hara2018,Piergiovanni2018EvolvingVideos,Tran2019VideoNetworks,Tran2018a,Xie2018} hinders the richness of final temporal information. The features before TGAP can be considered as features of different temporal regions of a clip or video. Although the receptive field might cover the whole clip, the effective receptive field has a Gaussian distribution \cite{Luo2017UnderstandingNetworks}, producing features focusing on different temporal regions of a clip. In order to discriminate actions, one part of the temporal feature might be more important than the others or the order of the temporal features might be more beneficial than simply averaging the temporal information. Therefore, TGAP ignores this ordering and fails to fully exploit the temporal information.

Therefore, we propose using the attention mechanism of BERT for better temporal modeling than TGAP. BERT determines which temporal features are more important with its multi-head attention mechanism.

To the best of our knowledge, our work is the first to propose replacing TGAP in 3D CNN architectures with late temporal modeling. We also consider that this study is the first to utilize BERT as a temporal pooling strategy in AR. We show that BERT performs better temporal pooling than average pooling, concatenation pooling, and standard LSTM. Moreover, we demonstrate that late temporal modeling with BERT improves the performances of various popular 3D CNN architectures for AR which are ResNeXt101, I3D, SlowFast, and R(2+1)D by using the split-1 of the HMDB51 dataset. Using BERT R(2+1)D architecture, we obtain the new state of the art results; 85.10\% and 98.69\% Top-1 performances in HMDB51 and UCF101 datasets, respectively. 

\section{Related Work on Action Recognition}
In this section, the AR literature is analyzed in two aspects: (i) temporal integration using pooling, fusion or recurrent architectures and (ii) 3D CNN architectures. 
\label{sec:related_work}
\subsection{Temporal Integration Using Pooling, Fusion or Recurrent Architectures}
\label{sec:intro_latePool}
Pooling is a well-known technique to combine various temporal features; concatenation, averaging, maximum, minimum, ROI, feature aggregation techniques and time-domain convolution are some of the possible pooling techniques \cite{Girdhar2017,Ng2015}.

Fusion frequently used for AR is very similar to pooling. Fusion is sometimes preferred instead of pooling in order to emphasize pooling location in the architecture or to differentiate information from different modalities. Late fusion, early fusion and slow fusion models on 2D CNN architectures can be performed by combining temporal information along the channel dimension at various points in CNN architectures \cite{Karpathy2014}. As a method, the two-stream fusion architecture in \cite{Feichtenhofer2016} creates spatio-temporal relationship with extra 3D convolution layer inserted towards the end of the architecture and fuses information from RGB and optical flow streams.

Recurrent networks are also commonly used for temporal integration. LSTMs are utilized for temporal (sequential) modeling on 2D CNN features extracted from the frames of a video \cite{Ng2015,Donahue2017a}. E.g., VideoLSTM \cite{Li2018} performs this kind of temporal modeling by using convolutional LSTM with spatial attention. RSTAN \cite{Du2018} implements both temporal and spatial attention concepts on LSTM and the attention weights of RGB and optical flow streams are fused. 

\subsection{3D CNN Architectures}
\label{sec:intro_3D}
3D CNNs are networks formed of 3D convolution throughout the whole architecture. In 3D convolution, filters are designed in 3D, and channels and temporal information are represented as different dimensions. Compared to the temporal fusion techniques, 3D CNNs process the temporal information hierarchically and throughout the whole network. Before 3D CNN architectures, temporal modeling was generally achieved by using an additional stream of optical flow or by using temporal pooling layers. However, these methods were restricted to 2D convolution and temporal information was put into the channel dimension. The downside of the 3D CNN architectures is that they require huge computational costs and memory demand compared to its 2D counterparts. 

The first 3D CNN for AR is the C3D model \cite{Tran2015}. Another successful implementation of 3D convolution is the Inception 3D model (I3D) \cite{Carreira2017}, in which 3D convolution is modeled in a much deeper fashion compared to C3D. The ResNet version of 3D convolution is introduced in \cite{Hara2018}. Then, R(2+1)D \cite{Tran2018a} and S3D \cite{Xie2018} architectures are introduced in which 3D spatio-temporal convolutions are factorized into spatial and temporal convolutions, and shown to be more effective than traditional 3D convolution architectures. Another important 3D CNN architecture is Channel-Separated Convolutional Networks (CSN) \cite{Tran2019VideoNetworks} which separates the channel interactions and spatio-temporal interactions which can be thought as the 3D CNN version of depth-wise separable convolution \cite{Howard2017}. 

Slow-fast networks \cite{Feichtenhofer2019SlowfastRecognition} can be considered as a joint implementation of both fusion techniques and 3D CNN architectures. There are two streams, namely fast and slow paths. Slow stream operates at a low frame and focuses on spatial information, as the RGB stream in traditional two-stream architectures, while fast stream operates at a high frame and focuses on temporal information as an optical flow stream in traditional two-stream architectures. There is information flow from the fast stream to the slow stream. 

Although 3D CNNs are powerful, they still lack an effective temporal fusion strategy at the end of the architecture.
\section{Proposed Methods}
In this part, the proposed methods of this study are introduced. Firstly, the main proposed method, namely BERT-based temporal modeling with 3D CNN for activity recognition, is presented in Section \ref{sec:proposed_method_BERT}. Next, some novel feature reduction blocks are proposed in Section \ref{sec:proposed_method_FR}. These blocks are utilized to reduce the number of parameters of the proposed BERT-based temporal modeling. Thirdly, the proposed BERT-based temporal modeling implementations on SlowFast architecture are examined in Section \ref{sec:proposed_method_bert_slowfast}. The reason to re-consider the BERT-based late temporal modeling on SlowFast architecture is due to its different two-stream structure from other 3D CNN architectures.

\subsection{BERT-based Temporal Modeling with 3D CNNs for 
Action Recognition}
\label{sec:proposed_method_BERT}

\begin{figure}
\centering
\includegraphics[width=1\linewidth]{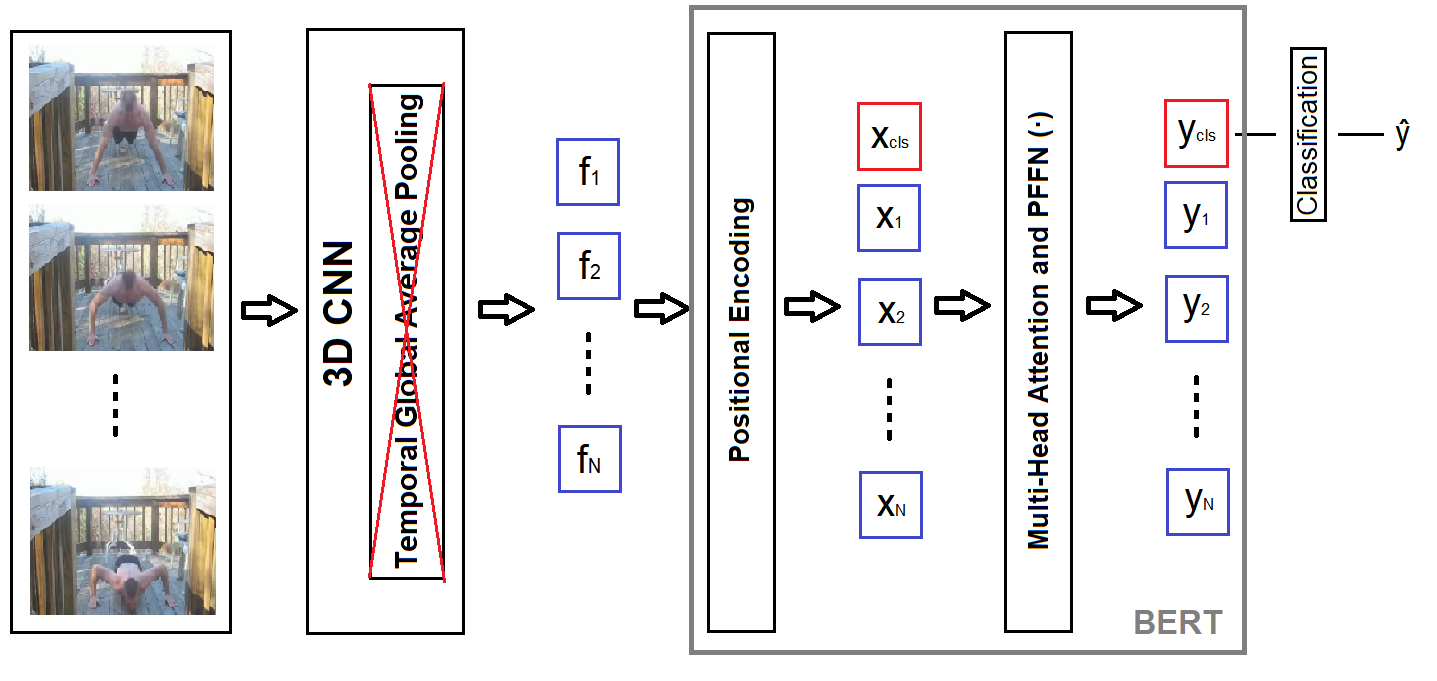}
\caption{BERT-based late temporal modeling}\label{fig:bert}
\end{figure}

Bi-directional Encoder Representations from Transformers (BERT) \cite{Devlin2018} is a bidirectional self-attention method, which has provided unprecedented success in many downstream Natural Language Processing (NLP) tasks. The bidirectional property enables BERT to fuse the contextual information from both directions, instead of relying upon only a single direction, as in former recurrent neural networks or other self-attention methods, such as Transformer \cite{Vaswani2017}. Moreover, BERT introduces challenging unsupervised pre-training tasks which leads to useful representations for many tasks.

Our architecture utilizes BERT-based temporal pooling as shown in Fig. \ref{fig:bert}. In this architecture, the selected $K$ frames from the input sequence are propagated through a 3D CNN architecture without applying temporal global average pooling at the end of the architecture. Then, in order to preserve the positional information, a learned positional encoding is added to the extracted features. In order to perform classification with BERT, additional classification  embedding ($\mathbf{x_{cls}}$) is appended as in \cite{Devlin2018} (represented as red box in Fig. \ref{fig:bert}). The classification of the architecture is implemented with the corresponding classification vector $\mathbf{y_{cls}}$ which is given to the fully connected layer, producing the predicted output label $\hat{y}$. 

The general single head self-attention model of BERT is formulated as:
\begin{equation}
    \mathbf{y_i} = PFFN \left( \frac{1}{N(x)} \sum_{\forall j} g(\mathbf{x_j})f(\mathbf{x_i},\mathbf{x_j}) \right),
\label{eq:bert_proposed}
\end{equation}
where $\mathbf{x_{i}}$ values are the embedding vectors that consists of extracted temporal visual information and its positional encoding;  $i$ indicates the index of the target output temporal position; $j$ denotes all possible combinations; and $N(x)$ is the normalization term. Function $g(\cdot)$ is the linear projection inside the self-attention mechanism of BERT, whereas  function $f(\cdot, \cdot)$ denotes the similarity between $\mathbf{x_i}$ and $\mathbf{x_j}$: $f(\mathbf{x_i},\mathbf{x_j}) = \textrm{softmax}_j(\theta(\mathbf{x_i})^T \phi(\mathbf{x_j}))$, where the functions $\theta(\cdot)$ and $\phi(\cdot)$ are also linear projections. The learnable functions $g(\cdot)$, $\theta(\cdot)$ and $\phi(\cdot)$ try to project the feature embedding vectors to a better space where the attention mechanism works more efficiently. The outputs of $g(\cdot)$, $\theta(\cdot)$ and $\phi(\cdot)$ functions are also defined as \textit{value}, \textit{query} and \textit{key}, respectively \cite{Vaswani2017}. $PFFN(\cdot)$ is Position-wise Feed-forward Network applied to all positions separately and identically: $PFFN(x) = \mathbf{W_2} GELU(\mathbf{W_1}\mathbf{x} + \mathbf{b1}) + \mathbf{b2}$, where $GELU(\cdot)$ is the Gaussian Error Linear Unit (GELU) activation function \cite{Hendrycks2016GaussianGELUs}.

The final decision of classification is performed with one more linear layer which takes $\mathbf{y_{cls}}$ as input. The explicit form of $\mathbf{y_{cls}}$ can be written as:
\begin{equation}
    \mathbf{y_{cls}} = PFFN \left( \frac{1}{N(x)} \sum_{\forall j} g(\mathbf{x_j})f(\mathbf{x_{cls}},\mathbf{x_j}) \right).
\end{equation}
Therefore, our use of the temporal attention mechanism for BERT is not only to learn the convenient subspace where the attention mechanism works efficiently but also to learn the classification embedding which learns how to attend the temporal features of the 3D CNN architecture properly.

A similar work for action recognition is implemented with non-local neural networks (NN) \cite{Wang2018a}. The main aim of non-local block is to create global spatio-temporal relations, since convolution operation is limited to local regions. For this aim, non-local blocks use a similar attention concept by using 1x1x1 CNN filters, in order to realize $g(\cdot)$, $\theta(\cdot)$ and $\phi(\cdot)$ functions. The main difference between the non-local and the proposed BERT attention is that non-local concept \cite{Wang2018a} is preferred to be utilized not at the end of the architecture, but some preferred locations inside the architecture. However, our BERT-based temporal pooling is implemented on the extracted features of the 3D CNN architecture and utilizes multi-head attention concept to create multiple relations with self-attention mechanism. Moreover, it utilizes positional encoding in order to preserve the order information and utilizes learnable classification token.

Another similar study for action recognition is the video action transformer network \cite{Girdhar2018} where the transformer is utilized in order to aggregate contextual information from other people and objects in the surrounding video. The video action transformer network deals with both action localization and action recognition; therefore, its problem formulation is different from ours and its attention mechanism needs to be reformulated for the late temporal modeling for action recognition. Differently from the video action transformer network, our proposed BERT-based late temporal modeling utilizes the learnable classification token, instead of using the pooled feature of the output of the backbone architecture.

\subsection{Proposed Feature Reduction Blocks: FRAB \& FRMB}
\label{sec:proposed_method_FR}

\begin{figure}[t]
	\centering
	\begin{subfigure}[b]{0.3\textwidth}{
    	\includegraphics[width=\linewidth]{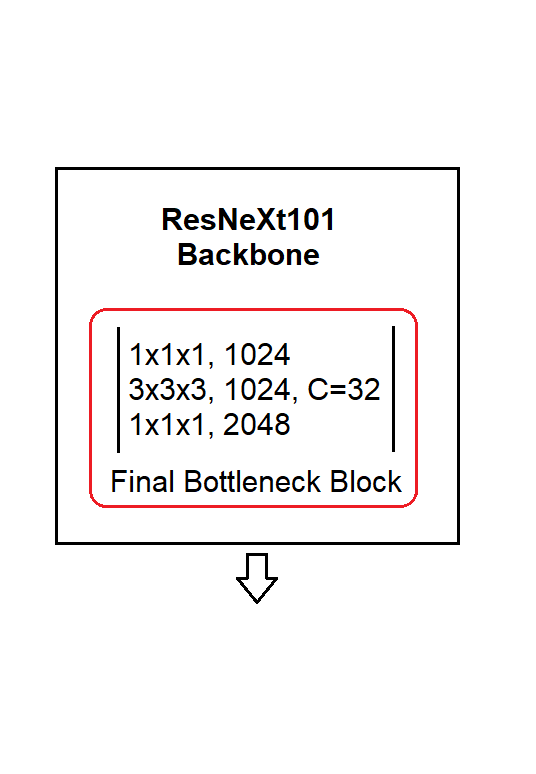}}
    \caption{Original}
	\end{subfigure}
	\begin{subfigure}[b]{0.3\textwidth}{
    	\includegraphics[width=\linewidth]{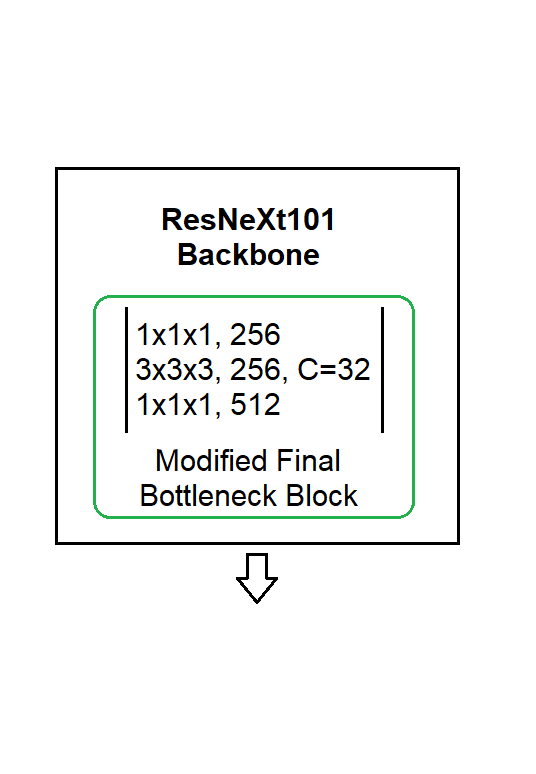}}
    \caption{FRMB}
	\end{subfigure}
	\begin{subfigure}[b]{0.3\textwidth}{
    	\includegraphics[width=\linewidth]{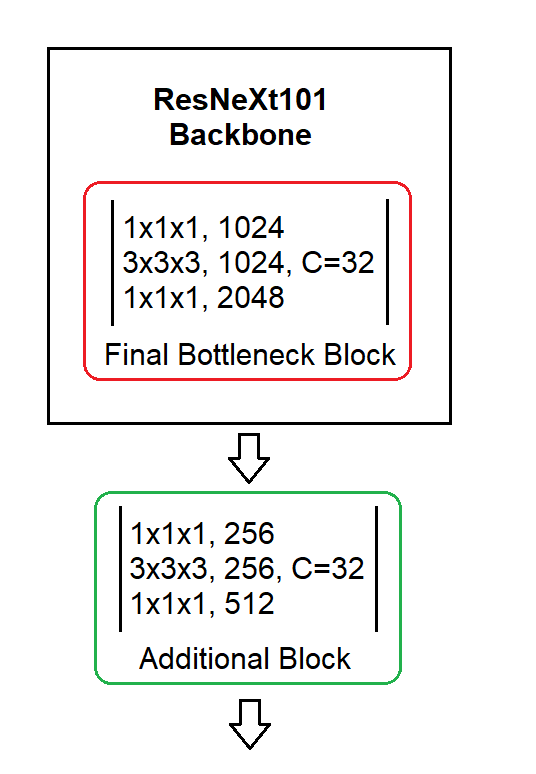}}
    \caption{FRAB}
	\end{subfigure}
	\caption{The implementations of Feature Reduction with Modified Block (FRMB) and Feature Reduction with Additional Block (FRAB)}
	\label{fig:feature_reduction}
\end{figure}

The computational complexity of BERT has a quadratic increase with the dimension of the output feature of the CNN backbone. As a result, if the dimension of the output feature is not reduced for specific backbones, the BERT module might have more parameters than the backbone itself. For instance, if the dimension of the output feature is 512, the single-layer BERT module has about 3 Million parameters, while the parameter size would be about 50 Million for the output feature dimension of 2048.

Therefore, in order to utilize BERT architecture in a more parameter efficient manner, two feature reduction blocks are proposed. These are Feature Reduction with Modified Block (FRMB) and Feature Reduction with Additional Block (FRAB). In FRMB, the final unit block of the CNN backbone is replaced with a novel unit block with the aim of feature dimension reduction. In FRAB, an additional unit block is appended to the backbone to reduce the dimension. An example implementation of FRMB and FRAB on ResNeXt101 backbone is presented in Figure \ref{fig:feature_reduction}.

The benefit of FRMB implementation is its better computational complexity and parameter efficiency over the FRAB implementation. Moreover, FRMB has even a better computational complexity and parameter efficiency than the original backbone. One possible downside of FRMB over FRAB is that the final block does not benefit from the pre-trained weights of the larger dataset if the feature reduction block is implemented only in the fine-tuning step but not in the pre-training. 

\subsection{Proposed BERT Implementations on SlowFast Architecture}
\label{sec:proposed_method_bert_slowfast}
SlowFast architecture \cite{Feichtenhofer2019SlowfastRecognition} introduces a different perspective for the two-stream architectures. Instead of utilizing two different modalities as two identical streams, the overall architecture includes two different streams (namely fast and slow streams or paths) with different capabilities for a single modality. In SlowFast architecture, the slow stream has a better spatial capability, while the fast stream has a better temporal capability. The fast stream has better temporal resolution and less channel capacity compared to the slow stream. 

\begin{figure}[t]
	\centering
	\begin{subfigure}[b]{0.8\textwidth}{
    	\includegraphics[width=\linewidth]{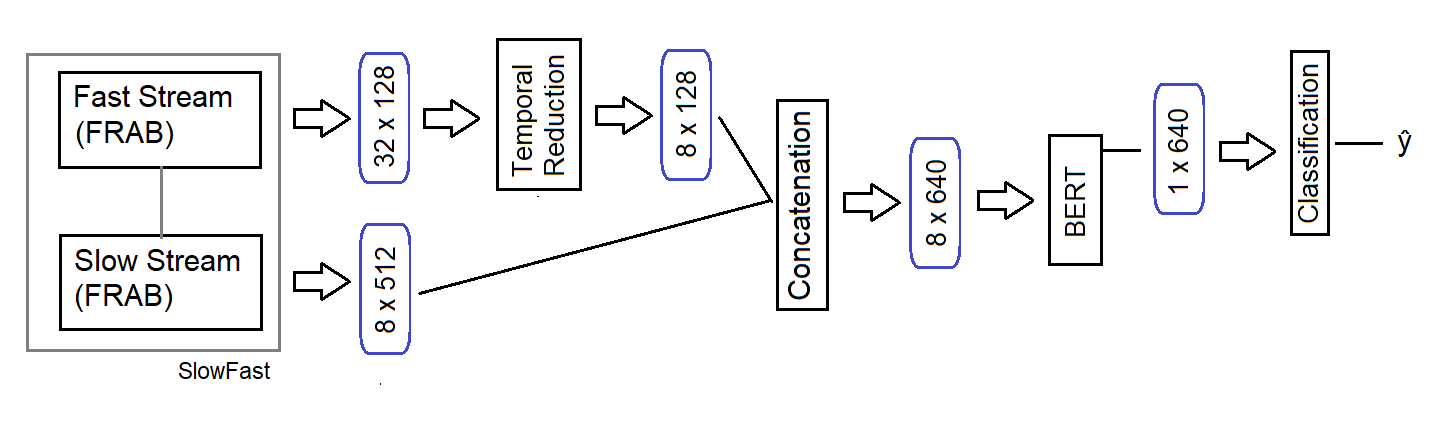}}
    \caption{Early-fusion}
	\end{subfigure}

	\begin{subfigure}[b]{0.8\textwidth}{
    	\includegraphics[width=\linewidth]{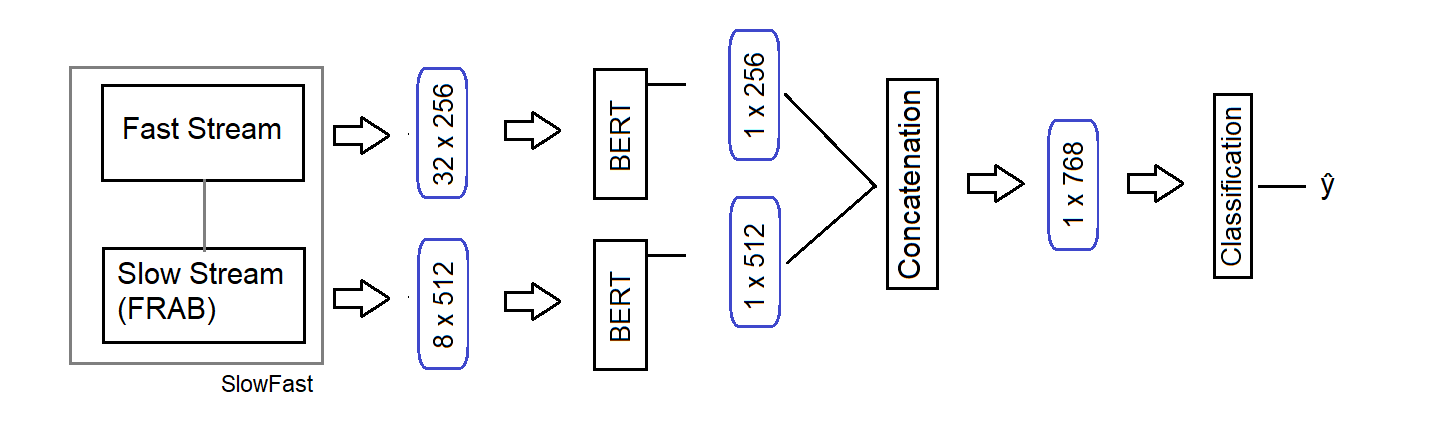}}
    \caption{Late-fusion}
	\end{subfigure}
	\caption{Early-fusion and late-fusion implementations of BERT on SlowFast architecture.}
	\label{fig:bert_slowfast}
\end{figure}

Due to its two-stream structure with different temporal resolutions, direct implementation of BERT-based late temporal modeling explained in Section \ref{sec:proposed_method_BERT} is not possible. Therefore, two alternative solutions are proposed in order to carry out BERT-based late temporal modeling on SlowFast architecture: Early-fusion BERT and late-fusion BERT. In early-fusion BERT, the temporal features are concatenated before the BERT layer and only a single BERT module is utilized. To make the concatenation feasible, the temporal resolution of the fast stream is decreased to the temporal resolution of the slow stream. In late-fusion BERT, two different BERT modules are utilized, one for each stream and the outputs of two BERT modules from two streams are concatenated. The figure for early-fusion and late-fusion is shown in Figure \ref{fig:bert_slowfast}.

\section{Experiments}
In this part, dataset, implementation details, ablation study, results on different architectures, and comparison with state-of-the-art sections are presented, respectively.
\subsection{Dataset}
\label{sec:dataset}

Four datasets are relevant for our study: HMDB51 \cite{Kuehne2011}, UCF101 \cite{Soomro2012}, Kinetics-400 \cite{Carreira2017} and IG65M \cite{Ghadiyaram2019Large-scaleRecognition} datasets. HMDB51 consists of $\sim$7k clips with 51 classes whereas UCF101 includes $\sim$13k clips with 101 classes. Both HMDB51 and UCF101 define three data splits and performances are calculated by averaging the results on these three splits. Kinetics-400 consists of about 240k clips with 400 classes. IG65M is a weakly supervised dataset which is collected by using the Kinetics-400 \cite{Carreira2017} class names as hashtags on Instagram. There are 65M clips from 400 classes. The dataset is not public for the time being but there are pre-trained models available. 

For analyzing the improvements of BERT on individual architectures (Section \ref{sec:experiments_different_architectures}), split 1 of the HMDB51 dataset is used, whereas the comparisons against the-state-of-the-art (See Section \ref{sec:experiments_state_of_the_art}) are performed by using the three splits of the HMDB51 and UCF101 datasets. Additionally, the ablation study (See Section \ref{sec:ablation_study}) is conducted using the three splits of HMDB51. Moreover, Kinetics-400 and IG65M are used for pre-trained weights of the architectures before fine-tuning on HMDB51 and UCF101. The pre-trained weights are obtained from the authors of architectures, which are ResNeXt, I3D, Slowfast, and R(2+1)D. Among these architectures, R(2+1)D is pre-trained with IG65M but the rest of the architectures are pre-trained with Kinetics-400.  

\subsection{Implementation Details}
For the standard architectures (with TGAP and without any modification to architectures), SGD with learning rate $10^{-2}$ is utilized, except I3D in which the learning rate is set to $10^{-1}$ empirically. For architectures with BERT, the ADAMW optimizer \cite{Loshchilov2017DecoupledRegularization} with learning rate $10^{-5}$ is utilized except I3D for which the learning rate is set to $10^{-4}$ empirically. For all training runs, the ``reducing learning rate on the plateau" schedule is followed. The data normalization schemes are selected conforming with the data normalization schemes of the pre-training of the architectures in order to benefit fully from pre-training weights. A multi-scale cropping scheme is applied for fine-tuning and testing of all architectures \cite{Wang2015}. In the test time, the scores of non-overlapping clips are averaged. The optical flow of the frames is extracted with the TV-L1 algorithm (Appendix A). 

In the BERT architecture, there are eight attention heads and one transformer block. The dropout ratio in $PFFN(\cdot)$ is set to 0.9. Mask operation is applied with 0.2 probability. Instead of using a mask token, the attention weight of the masked feature is set to zero. The classification token ($\mathbf{x_{cls}}$) and the learned positional embeddings are initialized as the zero-mean normal weight with 0.02 standard deviation. Default Torch linear layer initialization is used. Different from the I3D-BERT architecture, the linear layers of BERT are also initialized as the zero-mean normal weight with 0.02 standard deviation since it yields better results for I3D-BERT.  

\subsection{Ablation Study}
\label{sec:ablation_study}
In this section, we will analyze each step of our proposals and examine how our method compares with alternative pooling strategies (see Table \ref{table:ablation}). In this analysis, the ResNeXt101 backbone is utilized with the RGB modality, with a 112x112 input image size, and with 64-frame clips. In Table \ref{table:ablation}, temporal pool types, the existence of Feature Reduction with Modified Block (FRMB), the type of the optimizer, top1 performances, the number of parameters, and the number of operations are presented as the columns of the analysis.

\begin {table}[!t]
\centering
\caption{Ablation Study of RGB ResNeXt101 architecture for temporal pooling analysis on HMDB51. FRMB: Feature Reduction with Modified Block.}
\begin{tabular}{ | c | c | c | c | c | c | } 
 \hline
   \textbf{ Type of}  & \textbf{FRMB?} & \textbf{Optimizer} & \textbf{Top1}  & \textbf{\# of} & \textbf{\# of}\\
 \textbf{Temporal Pooling} & & & \textbf{ (\%)} & \textbf{Params} & \textbf{Operations} \\
 \hline \hline
  Average Pooling  &  & \multirow{2}{*}{SGD} & \multirow{2}{*}{74.46} & \multirow{2}{*}{47.63 M} & \multirow{2}{*}{38.56 GFlops} \\
  (Baseline) & & & & & \\
 \hline
  Average Pooling &  & ADAMW & 75.99 & 47.63 M & 38.56 GFlops\\ 
 \hline
  Average Pooling & \checkmark & ADAMW & 74.97 & 44.22 M & 38.36 GFlops\\ 
 \hline
  Concatenation  & \checkmark & ADAMW & 76.49 & 44.30 M& 38.36 GFlops \\
 \hline
  LSTM & \checkmark & ADAMW & 74.18 & 47.58 M & 38.37 GFlops\\ 
 \hline
  Concatenation +  & \multirow{2}{*}{\checkmark} & \multirow{2}{*}{ADAMW} & \multirow{2}{*}{76.84} & \multirow{2}{*}{47.45 M} & \multirow{2}{*}{38.36 GFlops} \\ 
  Fully Connected Layer &  & & & & \\
 \hline
  Non-local +  & \multirow{3}{*}{\checkmark} & \multirow{3}{*}{ADAMW} & \multirow{3}{*}{76.36} & \multirow{3}{*}{47.35 M}  & \multirow{3}{*}{38.43 GFlops} \\ 
  Concatenation +  &  & & & & \\
  Fully Connected Layer  &  & & & & \\
 \hline
  BERT pooling (Ours) & \checkmark & ADAMW & \textbf{77.49} & 47.38 M & 38.37 GFlops\\ 
 \hline
\end{tabular}
\label{table:ablation}
\end {table}

One important issue is the optimizer. For training BERT architectures in NLP tasks, the ADAM optimizer is usually selected \cite{Devlin2018}. However,  SGD is preferred for 3D CNN architectures \cite{Hara2018,Carreira2017,Feichtenhofer2019SlowfastRecognition,Tran2018a,Crasto2019MARS:Recognition}. Therefore, for training BERT, we select ADAMW (i.e. not ADAM), since ADAMW improves the generalization capability of ADAM \cite{Loshchilov2017DecoupledRegularization}. In this ablation study, ResNeXt101 architecture (with Average Pooling in Table \ref{table:ablation}) is also trained with ADAMW in Table \ref{table:ablation} which shows 1.5\% increase in performance compared to SGD.

In this ablation study, FRMB implementation is selected for two reasons over FRAB (see Section \ref{sec:proposed_method_FR} for FRAB and FRMB). Firstly, FRMB yields about 0.5\% better top1 performance than FRAB. Secondly, FRMB has better computational complexity and parameter efficiency than FRAB. From the experiments of the ablation study, it is observed that FRMB has lower computational complexity and better parameter efficiency at the cost of $\sim$1\% decrease in Top-1 performance compared to the standard backbone (Table \ref{table:ablation}).

For a fair comparison, we set the hyper-parameters of the other pooling strategies (LSTM, concatenation + fully connected layer, and non-local + concatenation + fully connected layer) such that the number of parameters and the number of operations of these temporal pooling strategies is almost the same compared to the proposed BERT pooling. LSTM is implemented in two stacks and with a hidden-layer size 450. The dimension of the inter-channel of a non-local attention block (the dimension size of the attention mechanism) is set equal to the input size to the non-local block which is 512. The number of nodes of a fully connected layer is determined according to the need for equal parameter size with the proposed BERT temporal pooling for a fair comparison.

When Table \ref{table:ablation} is analyzed, one can observe that among the six different alternatives (with FRMB), BERT has the best temporal pooling strategy. Additionally, the proposed FRMB-ResNeXt101-BERT provides 3\% better Top-1 accuracy than the ResNeXt101-Average Pooling (Baseline) despite the fact that FRMB-ResNeXt101-BERT has a better computational complexity and parameter efficiency than the ResNeXt101-Average Pooling (Baseline) (see Table \ref{table:ablation}). The BERT layer itself has about 3M parameters and negligible computational complexity with respect to the ResNeXt101 backbone. For the other temporal pooling strategies, LSTM worsens the performance with respect to the temporal average pooling. Concatenation and concatenation + fully connected layer are also other successful strategies in order to utilize the temporal features better than the average pooling. The addition of a non-local attention block before the concatenation + fully connected layer also decreases the performance compared to only concatenation + fully connected layer pooling implementation. It should be highlighted that the original implementation of the non-local study \cite{Wang2018a} also prefers not to utilize the non-local block at the end of the final three bottleneck blocks, which is a consistent fact with the experimental result of this study related with non-local implementation. 

\begin {table}[!t]
\centering
\caption{Ablation Study of BERT late temporal Modeling on HMDB51.}
\begin{tabular}{ | c | c | c | c |} 
 \hline
\text{Number of}  & \text{Number of} & \text{Learnable Classification Token} & \text{Top1}\\
\text{BERT Layers} & \text{Attention Heads} & \text{against Pooled Features} & \text{(\%)}\\
 \hline
 1 & 8 &  & 76.07 \\
 \hline
 1 & 1 & \checkmark & 76.97 \\
 \hline
 1 & 8 & \checkmark & \textbf{77.49} \\
 \hline
 2 & 8 & \checkmark & 77.24 \\
 \hline
\end{tabular}
\label{table:ablation2}
\end {table}

In addition, the ablation study of BERT late temporal modeling is performed and presented in Table \ref{table:ablation2}. These results examine the effects of the number of layers, the number of heads, and utilization of learnable classification token instead of the average pooled feature. Initially, the experiment of replacing the average of extracted temporal features with learnable classification token results in a 1.42\% Top-1 accuracy boost. Next, utilization of multi-head attention with eight attention heads improves the Top-1 performance of single-head attention with 0.52\%. Thirdly, increasing the number of layers from one to two worsens the top1 performance with 0.25\%. Moreover, the memory trade-off of every layer of BERT is about 3M. The reason behind the deterioration might be the fact that late temporal modeling is not as much complex as capturing rich linguistic information and a single layer might be enough to capture the temporal relationship between the output features of 3D CNN architectures.

\subsection{Results on Different 3D CNN Architectures}
\label{sec:experiments_different_architectures}
In this section, the improvements obtained by replacing TGAP with BERT pooling on popular 3D convolution architectures for action recognition is presented, including ResNeXt101 \cite{Hara2018}, I3D \cite{Carreira2017}, SlowFast\cite{Feichtenhofer2019SlowfastRecognition} and R(2+1)D \cite{Tran2018a}.

\subsubsection{ResNeXt Architecture}
ResNeXt architecture is essentially ResNet with group convolutions \cite{Hara2018}. For testing this architecture, the input size is selected as 112x112 as in the study of \cite{Hara2018,Crasto2019MARS:Recognition} and 64 frame length is utilized. 

\begin {table}[!t]
\centering
\caption{Analysis of ResNeXt101 architecture with and without BERT for RGB, Flow, and two-stream modalities on HMDB51 split-1}
\begin{tabular}{ | c | c | c | c | c | c | } 
 \hline
  \textbf{BERT} & \textbf{Modality} & \textbf{Top-1} & \textbf{\# Parameters} & \textbf{\# Operations}\\
 \hline \hline
  & RGB & 74.38 & 47.63 M & 38.56 GFlops\\ 
 \hline
  \checkmark& RGB & \textbf{77.25} & 47.38 M & 38.37 GFlops\\ 
 \hline \hline
  & Flow & 79.48 & 47.60 M & 34.16 GFlops\\ 
 \hline
  \checkmark& Flow & \textbf{82.03} & 47.36 M & 33.97 GFlops\\ 
 \hline \hline
  & Both & 82.09 & 95.23 M & 72.72 GFlops\\ 
 \hline
  \checkmark& Both & \textbf{83.99} & 94.74 M & 72.34 GFlops\\ 
 \hline
\end{tabular}
\label{table:resnext}
\end {table}

The results of the ResNeXt101 architecture is presented in Table \ref{table:resnext}. The performance of the architectures is compared over RGB modality, (optical) flow modality, and both (two-stream) in which both RGB and flow-streams are utilized, and the scores are summed from each stream. In this table, the number of parameters and operations of the architectures are also presented. For feature reduction, FRMB is chosen instead of FRAB and its reasoning is explained in Section \ref{sec:ablation_study}). Based on the results in Table \ref{table:resnext}, the most important observation is the improvement of the performance by using BERT over the standard architectures (without BERT) in \textit{all} modalities. 

\subsubsection{I3D Architecture}
I3D architecture is an Inception-type architecture. During I3D experiments, the input size is selected as 224x224 and 64 frame length is used conforming with the I3D study \cite{Carreira2017}. The result of BERT experiments on I3D architecture is given in Table \ref{table:I3D}. In this table, there are two BERT implementations that are with and without FRAB. For I3D-BERT architectures with FRAB, the final feature dimension of the I3D backbone is reduced from 1024 to 512 in order to utilize BERT in a more parameter efficient manner. However, contrary to the ResNeXt101-BERT architecture, FRAB is selected instead of FRMB, because FRAB obtains about 3.6\% better Top-1 result for RGB-I3D-BERT architecture on split1 of HMDB51.

The experimental results in Table \ref{table:I3D} indicate that BERT increases the performance of I3D architectures in all modalities. However, the increase in RGB modality is quite limited. For the Flow modality, although there is a performance improvement for BERT without FRAB, the implementation of BERT with FRAB performs worse than the standard I3D architecture, implying that preserving the feature size is more important for flow modality compared to RGB modality in I3D architecture. For the two-stream setting, both of the proposed BERT architectures perform equally with each other and perform better than standard I3D with 0.65\% Top-1 performance increase. Comparing with the ResNeXt101 architecture, the performance improvements brought by BERT temporal modeling is lower in I3D architecture. 

\begin {table}[!t]
\centering
\caption{The performance analysis of I3D architecture with and without BERT for RGB, Flow, and two-stream modalities on HMDB51 split-1}
\begin{tabular}{ | c | c | c | c | c | c |} 
 \hline
  \multirow{2}{*}{\textbf{BERT}} & \multirow{2}{*}{\textbf{Modality}} & \textbf{Feature} & \multirow{2}{*}{\text{Top-1}} & \multirow{2}{*}{\textbf{\# Parameters}} & \multirow{2}{*}{\textbf{\# Operations}}\\

  &  & \textbf{Reduction} & &  & \\ 
 \hline\hline
  & RGB & X & 75.42 & 12.34 M & 111.33 GFlops\\ 
 \hline
  \checkmark & RGB & FRAB &\textbf{75.75} & 16.40 M & 111.72 GFlops\\
 \hline
  \checkmark & RGB & X & 75.69 & 24.95 M & 111.44 GFlops\\ 
 \hline\hline
  & Flow &  X & 77.97 & 12.32 M & 102.52 GFlops\\ 
 \hline
  \checkmark & Flow & FRAB & 77.25 & 16.37 M & 102.91 GFlops\\ 
 \hline
  \checkmark & Flow & X &\textbf{78.37} & 24.92 M & 102.63 GFlops\\ 
 \hline\hline
  & Both &  X & 82.03 & 24.66 M & 213.85 GFlops\\ 
 \hline
  \checkmark & Both & FRAB &\textbf{82.68} & 32.77 M & 214.63 GFlops\\ 
 \hline
  \checkmark & Both & X &\textbf{82.68} & 49.87 M & 214.07 GFlops\\ 
 \hline
\end{tabular}
\label{table:I3D}
\end {table}

\subsubsection{SlowFast Architecture}
The SlowFast architecture in these experiments is derived from a ResNet-50 architecture. The channel capacity of the fast streams is one-eighth of the channel capacity of the slow stream. The temporal resolution of the fast stream is four times the temporal resolution for the slow stream. The input size is selected as 224x224 and 64-frame length is utilized with the SlowFast architecture conforming with the SlowFast study \cite{Feichtenhofer2019SlowfastRecognition}. Although it might be possible to utilize SlowFast architecture with also optical flow modality, the authors of SlowFast did not consider this strategy in their study. Therefore, in this effort, the analysis of BERT is also implemented by only considering the RGB modality. 

In order to utilize BERT architecture with fewer parameters, the final feature dimension of SlowFast backbone is reduced similar to the ResNeXt101-BERT and I3D-BERT architectures. Similar to the I3D-BERT architecture, FRAB is chosen instead of FRMB since FRAB obtains about 1.5\% better Top-1 result for SlowFast-BERT architecture on the split1 of HMDB51. For early-fusion BERT, the feature dimension of the slow stream is reduced from 2048 to 512 and the feature dimension of the fast stream is reduced from 256 to 128. For late-fusion BERT, only the feature dimension of the slow stream is reduced from 2048 to 512. The details about the size of the dimensions are presented in Figure \ref{fig:bert_slowfast}. The proposed implementation of BERT-based late temporal modeling on SlowFast architecture is presented in Section \ref{sec:proposed_method_bert_slowfast}.

\begin {table}[!t]
\centering
\caption{The performance analysis of SlowFast architecture with and without BERT for RGB modality on HMDB51 split-1}
\begin{tabular}{ | c | c | c | c | } 
 \hline
   \textbf{BERT} & \textbf{Top-1} & \textbf{\# Parameters} & \textbf{\# Operations} \\
 \hline
   & 79.41 & 33.76 M & 50.72 GFlops\\ 
 \hline
  \checkmark (\textit{early-fusion})  & 79.54 &43.17 M & 52.39 GFlops\\ 
 \hline
  \checkmark (\textit{late-fusion})  & \textbf{80.78} & 42.04 M & 52.14 GFlops\\ 
 \hline
\end{tabular}
\label{table:SlowFast}
\end {table}

The results for BERT on SlowFast architecture are given in Table \ref{table:SlowFast}. First of all, both BERT solutions perform better than the standard SlowFast architecture but the improvement of early-fusion method is quite limited. Late-fusion BERT improves the top1 performance of standard SlowFast architecture with about 1.3 \%. From the number of parameters perspective, the implementation of BERT on SlowFast architecture is not as much as efficient in comparison to ResNeXt101 architecture because of the FRAB implementation instead of FRMB as in the case of I3D-BERT. Moreover, the parameter increase of RGB-SlowFast-BERT is even higher than RGB-I3D-BERT because of the two-stream implementation of SlowFast network for RGB input modality. The increase in the number of operations is also higher in the implementation of SlowFast-BERT than the I3D-BERT and ResNeXt101-BERT because of the higher temporal resolution in SlowFast architecture and two-stream implementation for RGB modality. 

For the two alternatives proposed BERT solution in Table \ref{table:SlowFast}, late-fusion BERT yields better performance with better computational complexity in contrast with early-fusion BERT. Although the attention mechanism is implemented jointly on the concatenated features, the destruction of the temporal richness of fast stream to some degree might be the reason for the inferior performance of the early-fusion BERT.

\subsubsection{R(2+1)D Architecture}
R(2+1)D \cite{Tran2018a} architecture is a ResNet-type architecture consisting of separable 3D convolutions in which temporal and spatial convolutions are implemented separately. For this architecture, 112x112 input dimensions are applied following the paper, and 32-frame length is applied instead of 64-frame because of the huge memory demand of this architecture and to be consistent with the paper \cite{Tran2018a}. The selected R(2+1)D architecture has 34 layers and implemented with basic block type instead of bottleneck block type (for further details about block types, see \cite{Hara2018}). The most important difference of R(2+1)D experiments from the previous architectures is the utilization of the IG65M pre-trained weights, instead of Kinetics pre-trained weights (see Section \ref{sec:dataset} for details). Therefore, this information should always be considered while comparing this architecture with the aforementioned ones. The analysis of R(2+1)D BERT architecture is limited to RGB modality, since the study \cite{Ghadiyaram2019Large-scaleRecognition} of the IG65M dataset where R(2+1)D architecture is preferred is limited to RGB modality.   

\begin {table}[!t]
\centering
\caption{The performance analysis of R(2+1)D architecture with and without BERT for RGB modality on HMDB51 split-1}
\begin{tabular}{ | c | c | c | c | } 
 \hline
  \textbf{BERT} & \textbf{Top-1} & \textbf{\# Parameters} & \textbf{\# Operations} \\
 \hline
   & 82.81 & 63.67 M & 152.95 GFlops\\ 
 \hline
  \checkmark & \textbf{84.77} & 66.67 M & 152.97 GFlops\\ 
 \hline
\end{tabular}
\label{table:R(2+1)D}
\end {table}

The experiments for BERT on R(2+1)D architecture are presented in Table \ref{table:R(2+1)D}. The feature dimension of R(2+1)D architecture is already 512 which is the same with the reduced feature dimension of ResNeXt101 and I3D backbones for BERT implementations. Therefore, we do not use FRMB or FRAB for R(2+1)D. There is an increase of about 3M parameters and the increase in the number of operations is still negligible. The performance increase of BERT on R(2+1)D architecture is about 2\% which is a significant increase for RGB modality as in the case of ResNeXt101-BERT architecture. 

\subsection{Comparison with State-of-the-Art}
\label{sec:experiments_state_of_the_art}
In this section, the results of the best BERT architectures from the previous section are compared against the state-of-the-art methods. For this aim, two leading BERT architectures are selected among all the test methods: Two-Stream BERT ResNeXt101 and RGB BERT R(2+1)D (see Section \ref{sec:experiments_different_architectures}). Note that these two architectures use different pre-training datasets, namely IG65 and Kinetics-400 for ResNext101 and R(2+1)D, respectively. 

The results of the architectures on HMDB51 and UCF101 datasets are presented in Table \ref{table:literature_results}. The table indicates if an architecture employs explicit optical flow. Moreover, the table lists the pre-training dataset used by the methods.

As shown in Table \ref{table:literature_results}, BERT increases the Top-1 performance of the two-stream ResNeXt101 with 1.77\% and 0.41\% in HMDB51 and UCF101, respectively. Additionally, BERT improves the Top-1 performance of RGB R(2+1)D (32f) with 3.5 \% and 0.48\% in HMDB51 and UCF101, respectively, where 32f corresponds to 32-frame length. The results obtained by the R(2+1)D BERT (64f) architecture pre-trained with the IG65M dataset is the current state-of-the-art result in AR for HMDB51 and UCF101, to the best of our knowledge. 

Among the architectures pre-trained in Kinetics-400, the two-stream ResNeXt101 BERT is again the best in HMDB51 but the second-best in the UCF101 dataset. This might be owing to the fact that HMDB51 involves some actions that can be resolved only using temporal reasoning and therefore benefits from BERT's capacity.   

An important point to note from the table is the effect of pre-training with the IG65M dataset. RGB R(2+1)D (32f) (without Flow) pre-trained with IG65M obtains about 6\% and 1.4\% better Top-1 performance in HMDB51 and UCF101, respectively than the one pre-trained with Kinetics-400, indicating the importance of the number of samples in the pre-training dataset even if the samples are collected in a weakly-supervised manner.

\begin {table}[!t]
\centering
\caption{Comparison with the state-of-the-art.}
\begin{tabular}{ | c | c | c | c | c | } 
 \hline
  \textbf{} & \textbf{Uses} & \textbf{Extra } & \textbf{} & \textbf{} \\ 
  \textbf{Model} & \textbf{Flow?} & \textbf{Training Data} & \textbf{HMDB51} & \textbf{UCF101} \\ 
 \hline\hline
  IDT \cite{Wang2013a} & \checkmark &  & 61.70 & - \\ 
 \hline
  Two-Stream \cite{Simonyan2014} & \checkmark & ImageNet & 59.40 & 88.00\\
 \hline
  Two-stream Fusion + IDT \cite{Feichtenhofer2016} & \checkmark & ImageNet & 69.20 & 93.50 \\ 
 \hline
  ActionVlad + IDT \cite{Girdhar2017} & \checkmark & ImageNet & 69.80 & 93.60\\ 
 \hline
  TSN \cite{Wang2018} & \checkmark & ImageNet & 71.00 & 94.90\\ 
 \hline
  RSTAN + IDT \cite{Du2018}  &  \checkmark & ImageNet & 79.90 & 95.10\\ 
 \hline
  TSM \cite{Lin2018TSM:Understanding} &  & Kinetics-400 & 73.50 & 95.90\\ 
 \hline
  R(2+1)D \cite{Tran2018a} &  & Kinetics-400 & 74.50 & 96.80\\
 \hline
  R(2+1)D \cite{Tran2018a} & \checkmark & Kinetics-400 & 78.70 & 97.30\\
 \hline
  I3D \cite{Carreira2017} & \checkmark & Kinetics-400 & 80.90 & 97.80 \\
 \hline
  MARS + RGB + Flow \cite{Crasto2019MARS:Recognition} & \checkmark & Kinetics-400 & 80.90 & \textbf{98.10}\\ 
 \hline
  FcF \cite{Piergiovanni2018RepresentationRecognition} &  & Kinetics-400 & 81.10 & - \\ 
 \hline
  ResNeXt101 &  \checkmark &  Kinetics-400 & 81.78 & 97.46 \\
 \hline
  EvaNet \cite{Piergiovanni2018EvolvingVideos} & \checkmark & Kinetics-400 & 82.3 & - \\ 
 \hline
  HAF+BoW/FV halluc \cite{Wang2019HallucinatingCNNs} &  &  Kinetics-400 & 82.48 & - \\
 \hline
  ResNeXt101 BERT \textbf{(Ours)} &  \checkmark &  Kinetics-400 & \textbf{83.55} & 97.87\\
 \hline
  R(2+1)D (32f) &  &  IG65M & 80.54 & 98.17\\ 
 \hline
  R(2+1)D BERT (32f) \textbf{(Ours)} &  &  IG65M & 83.99 & 98.65\\
 \hline
  R(2+1)D BERT (64f) \textbf{(Ours)} &  &  IG65M & \textbf{85.10} & \textbf{98.69}\\
 \hline
\end{tabular}
\label{table:literature_results}
\end {table}

\section{Conclusions}
\label{sec:conclusions}
This study combines the two major components from AR literature, namely late temporal modeling and 3D convolution. Although there are many pooling, fusion, and recurrent modeling strategies that are applied to the features from 2D CNN architectures, we firmly believe that this manuscript is the first study that removes temporal global average pooling (TGAP) and better employs temporal information at the output of 3D CNN architectures. To utilize these temporal features, an attention-based mechanism namely BERT is selected. The effectiveness of this idea is proven on most of the popular 3D CNN architectures which are ResNeXt, I3D, SlowFast, and R(2+1)D. In addition, significant improvements over the-state-of-the-art techniques are obtained in HMDB51 and UCF101 datasets. 

The most important contribution of this study is the introduction of the late temporal pooling concept, paving the way for better late temporal pooling strategies over BERT on 3D CNN architectures as future work, although better performance is obtained with BERT over average pooling, concatenation pooling, and standard LSTM pooling. A possible research direction might be proposals for parameter efficient BERT implementations that do not need feature reduction blocks (FRMB or FRAB) which decreases the capabilities of the final extracted features because of the reduction in the dimension of features. Additionally, as future work, unsupervised concepts can still be proposed on BERT 3D CNN architectures, since the real benefits of BERT architecture rise to the surface with unsupervised techniques. Finally, the proposed method has also the potential to improve similar tasks with AR, such as temporal and spatial action localization and video captioning.

\begin{comment}

\clearpage\mbox{}Page \thepage\ of the manuscript.
\clearpage\mbox{}Page \thepage\ of the manuscript.

This is the last page of the manuscript.
\par\vfill\par
Now we have reached the maximum size of the ECCV 2020 submission (excluding references).
References should start immediately after the main text, but can continue on p.15 if needed.
\end{comment}

\section*{ACKNOWLEDGMENTS}
This work was supported by an Institutional Links grant under the Newton-Katip Celebi partnership, Grant No. 217M519 by the Scientific and Technological Research Council of Turkey (TUBITAK) and ID [352335596] by British Council, UK. The numerical calculations reported in this paper were partially performed at TUBITAK ULAKBIM, High Performance and Grid Computing Center (TRUBA resources)

\clearpage
\bibliographystyle{splncs04}
\bibliography{egbib}
\end{document}